\documentclass[letterpaper]{article}

\usepackage{microtype}
\usepackage{graphicx}
\usepackage{booktabs} 

\usepackage{hyperref}

\usepackage[utf8]{inputenc} 
\usepackage[T1]{fontenc}    
\usepackage{url}            
\usepackage{amsfonts}       
\usepackage{nicefrac}       
\usepackage{microtype}      
\usepackage{xcolor}         

\usepackage{caption}
\usepackage{subcaption}
\usepackage{mathtools}
\usepackage{tabularx}

\usepackage{aaai_style/aaai}
\usepackage{times}
\usepackage{helvet}
\usepackage{courier}
\usepackage{amsmath}
\usepackage{amssymb}
\usepackage{mathtools}
\usepackage{amsthm}

\theoremstyle{plain}

\theoremstyle{definition}

\theoremstyle{remark}

\begin{document}
\title{Self-Imitation Learning from Demonstrations}
\author{%
  George Pshikhachev\\
  JetBrains Research\\
  \texttt{georgii39@gmail.com}
  \And
  Dmitry Ivanov\\
  JetBrains Research\\
  \texttt{dimonenka@mail.ru}\\
  \AND
  Vladimir Egorov\\
  JetBrains Research\\
  \texttt{vladimirrim98@gmail.com}
  \And
  Aleksei Shpilman\\
  JetBrains Research\\
  \texttt{alexey@shpilman.com}
}
\maketitle
\begin{abstract}
Despite the numerous breakthroughs achieved with Reinforcement Learning (RL), solving environments with sparse rewards remains a challenging task that requires sophisticated exploration. Learning from Demonstrations (LfD) remedies this issue by guiding the agent's exploration towards states experienced by an expert. Naturally, the benefits of this approach hinge on the quality of demonstrations, which are rarely optimal in realistic scenarios. Modern LfD algorithms require meticulous tuning of hyperparameters that control the influence of demonstrations and, as we show in the paper, struggle with learning from suboptimal demonstrations. To address these issues, we extend Self-Imitation Learning (SIL), a recent RL algorithm that exploits the agent's past good experience, to the LfD setup by initializing its replay buffer with demonstrations. We denote our algorithm as SIL from Demonstrations (SILfD). We empirically show that SILfD can learn from demonstrations that are noisy or far from optimal and can automatically adjust the influence of demonstrations throughout the training without additional hyperparameters or handcrafted schedules. We also find SILfD superior to the existing state-of-the-art LfD algorithms in sparse environments, especially when demonstrations are highly suboptimal.
\end{abstract}

\section{Introduction}

\begin{figure*}[t]
\centering
\begin{subfigure}{.45\linewidth}
\centering
\includegraphics[width=\linewidth]{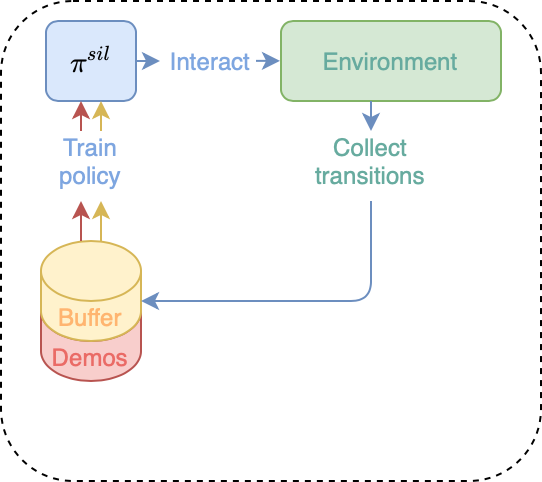}
\caption{SILfD -- for demonstrations with rewards}
\label{fig:scheme_a}
\end{subfigure}\hfill%
\begin{subfigure}{.45\linewidth}
\includegraphics[width=\linewidth]{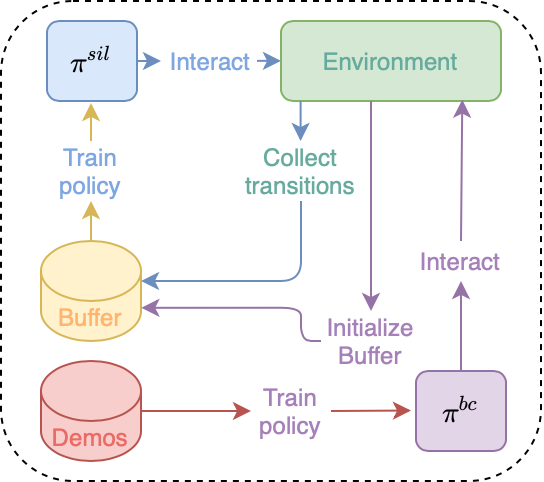}
\caption{SILfBC -- for demonstrations without rewards}
\label{fig:scheme_b}
\end{subfigure}
\caption{Schematic architectures of the proposed algorithms. The agent is represented by the policy $\pi^{\text{sil}}$ trained with Self-Imitation Learning. a) SILfD. The replay buffer that stores the agent experience is initialized with a set of expert demonstrations. The agent learns from both its own experience and the demonstrations. b) SILfBC. First, an auxiliary policy $\pi^{\text{bc}}$ is trained with Behavioural Cloning to mimic the expert based on a set of demonstrations. Then, the replay buffer is initialized with the experience of the auxiliary policy. The agent learns from both its own experience and the experience of the auxiliary policy.}
\label{fig:scheme}
\end{figure*}

Deep Reinforcement Learning (RL) algorithms have recently achieved multiple breakthroughs in solving games \cite{mnih2015human,moravvcik2017deepstack,berner2019dota,quake,brown2020combining}, hard visuomotor \cite{levine2016end} and manipulation \cite{gu2017deep} tasks, but some of these algorithms additionally rely on incorporating information from human demonstrations \cite{silver2016mastering,vinyals2019grandmaster}. Finding the optimal solution requires tremendous amount of environment interactions, which makes RL algorithms costly and dependent on sophisticated exploration techniques. This is especially true for environments with sparse rewards where encountering a positive reward requires a long and precise sequence of actions. An alternative approach is to additionally leverage a set of expert demonstrations, which has shown to help with exploration difficulties and provide magnitudes of improvement in learning speed and performance. This approach is known in the literature as Learning from Demonstrations (LfD) \cite{atkeson1997robot,schaal1997learning}. 

LfD algorithms can be attributed to one of the three categories based on the technique to incorporate demonstrations. The first technique is to treat demonstrations as additional learning references by placing them in the experience replay buffer \cite{hester2018deep,vecerik2017leveraging,gao2018reinforcement,nair2018overcoming,paine2019making}. The second technique is to optimize a mixture of reinforcement and imitation objectives by introducing either additional rewards \cite{kang2018policy,zhu2018reinforcement,zolna2019reinforced,brys2015reinforcement,hussenot2020show}, loss terms \cite{hester2018deep,rajeswaran2017learning,nair2018overcoming}, or hard constraints \cite{jing2020trpofd}. The third technique is to initialize agent's parameters with supervised \cite{silver2016mastering,rajeswaran2017learning,scheller2020sample} or imitation \cite{cheng2018fast} pretraining. Despite the impressive results on a variety of problems, modern algorithms typically assume access to high-quality demonstrations and, as our experiments confirm, degrade when demonstrations are noisy or suboptimal. Furthermore, these algorithms rely on additional techniques and hyperparameters to properly balance between learning from agent and expert experience.


Self-Imitation Learning (SIL) \cite{oh2018self} is a recent RL algorithm that imitates agent's past positive experience while ignoring negative experience. SIL has shown to fit particularly well in environments with sparse rewards where it can mimic complex behaviour required to reach the reward. Still, encountering the reward in the first place can be problematic, especially when using naive exploration. In this paper we show that SIL can greatly benefit from expert demonstrations by alleviating the need to encounter positive experience and propose Self-Imitation Learning from Demonstrations (SILfD).

The idea behind SILfD is to initialize the experience replay buffer with demonstrations. While similar ideas are used in algorithms like DQfD \cite{hester2018deep} and DDPGfD \cite{vecerik2017leveraging}, we argue that SIL is a natural choice for the LfD setting. The focus on positive experience and the prioritization mechanism ensure that SILfD selects the most useful demonstrations if their quality varies, forgoes learning from suboptimal demonstrations when they become obsolete, and dynamically balances between learning from agent and expert experience based on its current usefulness. Furthermore, SILfD does not introduce any new hyperparameters related to demonstrations and can learn from as few as one demonstration. Additionally, we propose SILfBC, an extension of SILfD to the cases where the rewards are not observed in demonstrations, which can be especially relevant when demonstrations are collected by a human expert.

We compare SILfD with the existing LfD algorithms in several environments: a toy hard-exploration environment Chain \cite{strens2000bayesian}; four DeepMind Control Suite tasks \cite{tassa2018deepmind} with continuous actions and sparsified rewards; and Pommerman environment  \cite{resnick2018pommerman} with procedural map generation. Experiments show that both SILfD and SILfBC outperform the existing state-of-the-art LfD algorithms, especially when demonstrations are highly suboptimal.

\section{Background and Notations}

\subsection{Reinforcement Learning} 


We consider the standard Markov Decision Process $\langle S, A, r, T, \gamma \rangle$, where 

\begin{itemize}
\item $S$ denotes the space of states $s$, $A$ denotes the space of actions $a$,
\item $r : S \times A \rightarrow \mathbb{R}$ denotes the reward function,
\item $T : S \times A \rightarrow \Delta(S)$ denotes the transition function, where $\Delta$ denotes probability distribution,
\item $\gamma \in (0, 1)$ denotes the discount factor.
\item Further, $R_t = \sum_{n=t}^{\infty}\gamma^{n-t}r_n$ denotes return, where subscripts $t$ and $n$ denote time steps,
\item $\pi_\theta : S \rightarrow \Delta(A)$ denotes policy parameterized by $\theta$,
\item $V(s) = \mathbb{E}_{\pi_\theta} \left[ R_t \vert s_t = s \right]$, $Q(s, a) = \mathbb{E}_{\pi_\theta} \left[ R_t \vert s_t = s, a_t = a \right]$, and $A(s, a) = Q(s, a) - V(s)$ respectively denote value, Q-value, and advantage functions.
\end{itemize}

Advantage-Actor-Critic (A2C) \cite{a3c} is one of the most prevalent RL frameworks where the Actor chooses actions in the environment by predicting policy in a given state while the Critic evaluates the state to aid the Actor's learning. Proximal Policy Optimization (PPO) \cite{ppo} is based on the A2C framework and focuses on staying within a trust region during the updates of policy parameters.




\subsection{Imitation Learning}

The purpose of Imitation Learning (IL) is to train a policy that mimics expert behaviour, the samples of which are stored in a buffer of demonstrations $\mathcal{D} = ({s, a})$. Generative Adversarial Imitation Learning (GAIL) is a recent algorithm that trains two adversarial models: discriminator and generator \cite{ho2016generative}. The discriminator is a binary classifier that distinguishes the generated and the expert transitions, whereas the generator constitutes the policy and tries to confuse the discriminator. The version of GAIL with a weighted objective denoted as wGAIL is the state-of-the-art in IL from suboptimal demonstrations \cite{wang2021wgail}. This algorithm is based on an observation that in the case of demonstrations being of diverse quality, the better demonstrations, which tend to be more consistent than the potentially noisy suboptimal behaviour, should be weighted higher. This relative consistency can be measured as the confidence of the discriminator' predictions. 

Similarly to IL, the offline RL aims to train an agent given a sample of experience, but prohibits interactions with the environment. Behavioral Cloning (BC) is a classic offline RL algorithm that trains a policy to predict the demonstrated action for a given state by maximizing log-likelihood \cite{pomerleau1991efficient}. Decision Transformer (DT) can be seen as a modern analogue to BC \cite{chen2021decision} This model treats a projection of a past state-action pair and a desired return as a token. A casually masked sequence of such tokens representing past trajectory is passed through several attention layers and a linear decoder to predict an action that leads to the desired return in a given state. DT performs at least comparably with offline TD-based algorithms and BC and is able to extrapolate to returns beyond those provided during training.

\subsection{Learning from Demonstrations}

Unlike IL, Learning from Demonstrations (LfD) assumes both the reward signal $r$ and a buffer of demonstrations $\mathcal{D} = ({s, a, r})$ to be available. Typically, LfD algorithms use demonstrations to increase sample efficiency and aid exploration in environments with sparse rewards. \cite{hester2018deep} propose DQfD which extends DQN \cite{mnih2015human} to the LfD setup by initializing the replay buffer with demonstrations, providing demonstrations with a priority bonus, pretraining Q-network offline, mixing 1-step and n-step losses, and regularizing the network with an auxiliary supervised loss. As an analogue of DQfD for continuous control, \cite{vecerik2017leveraging} propose DDPGfD by applying similar modifications to the critic of DDPG. As an alternative approach, POfD \cite{kang2018policy} enforces occupancy measure matching between the agent and the expert by shaping the reward with the predictions of a GAIL-like discriminator. Similar to POfD ideas are employed in \cite{zhu2018reinforcement,zolna2019reinforced}. The state-of-the-art in LfD is the unnamed method from \cite{jing2020trpofd} that we denote as TRPOfD due to it being based on Trust-Region Policy Optimization \cite{schulman2015trust}. TRPOfD takes a similar to POfD route of guided exploration, but instead of optimizing a mixed reward it introduces a hard constraint on the divergence from the expert policy that is relaxed overtime. Finally, supervised pretraining from demonstrations with BC is routinely used to assist solving complex tasks, e.g. Go \cite{silver2016mastering} and Minecraft \cite{scheller2020sample}.

\subsection{Self-Imitation Learning}

SIL \cite{oh2018self} aims to reproduce agent's past good decisions based on the experience stored in a replay buffer $\mathcal{B} = ({s, a, r})$. The algorithm alternates between the standard on-policy update of A2C or PPO and the off-policy update that minimizes A2C loss with clipped advantages:

\begin{equation}
  L_{\text{policy}}^{\text{sil}} = - \mathbb{E}_{\mathcal{B}} [ \log \pi_\theta A^+_\phi(s, a) ] - \alpha \mathcal{H}(\pi_\theta)\label{eq_sil_policy}
\end{equation}
\begin{equation}
  L_{\text{value}}^{\text{sil}} = \mathbb{E}_{\mathcal{B}} [A^+_\phi(s, a)]^2\label{eq_sil_value}
\end{equation}

where $\theta$ and $\phi$ are the parameters of the Actor and the Critic, $(\cdot)^+ = \max(\cdot, 0)$ ensures that only good transitions are considered for updates, advantage is estimated as $A_\phi(s, a) = R - V_\phi(s)$, $\mathcal{H}(\pi_\theta) = \mathbb{E}_{\pi_\theta}[-\log{\pi_\theta(a|s)}]$ denotes the entropy of the policy, and $\alpha \geq 0$.


\cite{oh2018self} theoretically justify SIL by connecting it to the lower-bound soft Q-learning, an algorithm that approximates the lower bound of the optimal soft Q-value $Q^*$ by minimizing:

\begin{equation}\label{eq_sil_lb}
    L^{\text{lb}} = \mathbb{E}_{\mathcal{B}}[(R^\mu - Q_\phi(s, a))^+]^2
\end{equation}

where $R_t^\mu = r_t + \sum_{n=t+1}^{\infty}\gamma^{n-t}(r_n + \alpha \mathcal{H}_n(\mu))$ is the entropy-regularized return. Specifically, the authors show that minimizing $L^{\text{lb}}$ is equivalent to minimizing $L_{\text{policy}}^{\text{sil}}$ and $L_{\text{value}}^{\text{sil}}$ when $\alpha \rightarrow 0$. To improve training efficiency, SIL uses prioritized replay buffer \cite{schaul2016prioritized}.

\section{Self-Imitation Learning from Demonstrations}\label{section_our}

While SIL shines in exploiting past good experience, encountering such experience can be problematic, especially in sparse environments. An alternative source of good experience can be a set of expert demonstrations $\mathcal{D}$. As an extension of SIL to the LfD setting, we propose SIL from Demonstrations (SILfD) based on one simple modification of the original SIL: the experience replay buffer is initialized with demonstrations $\mathcal{D}$ that are preserved in the buffer throughout the training  (Fig. \ref{fig:scheme_a}). The policy and value losses are accordingly modified:  

\begin{equation}
  L_{\text{policy}}^{\text{sil}} = - \mathbb{E}_{\mathcal{B} \cup \mathcal{D}} [ \log \pi_\theta A^+_\phi(s, a) ] - \alpha \mathcal{H}(\pi_\theta)\label{eq_our_policy}
\end{equation}
\begin{equation}
  L_{\text{value}}^{\text{sil}} = \mathbb{E}_{\mathcal{B} \cup \mathcal{D}} [A^+_\phi(s, a)]^2\label{eq_our_value}
\end{equation}

\subsection{Properties of SILfD}\label{section_our_properties}

Despite its simplicity, the proposed method has several properties that are desirable from LfD algorithms. Specifically, SILfD discerns useful experience in demonstrations and automatically adjusts the effect of demonstrations as the agent improves, which is crucial for dealing with noise in the demonstrations and for outperforming the expert. Below we elaborate on the origin and the implications of these properties in SILfD and discuss whether the existing algorithms have these properties. We experimentally verify these properties of SILfD in Section \ref{section_experiments}.

\textbf{Discerning useful demonstrations.} In order to exploit demonstrations, the existing LfD algorithms rely on additional techniques and hyperparameters that control their influence. The examples are the priority bonus for demonstrations in \cite{hester2018deep,vecerik2017leveraging,paine2019making}, the supervised loss term in \cite{hester2018deep}, the reward for imitation in \cite{rajeswaran2017learning,kang2018policy,zhu2018reinforcement,hussenot2020show}, the hard constraint on imitation in \cite{jing2020trpofd}. However, this approach forces the agent to balance the original and the imitation goals and may prevent the agent from reaching the optimal policy. A particular scenario where this approach may fail is the setting of noisy demonstrations, i.e. when the demonstrated behaviour is inconsistent and of varied quality. Overdependence on such demonstrations without discerning the useful experience may hinder the final performance.

Our algorithm is different from the existing LfD algorithms in that it treats demonstrations as additional learning references identical to the agent's own experience and does not require additional hyperparameters to control the influence of the demonstrations. In the setting of noisy demonstrations, the potentially useless experience (that corresponds to negative advantages) has null contribution to the loss function (\ref{eq_our_policy}) used in SILfD, while the prioritization mechanism ensures that such demonstrations are not sampled from the buffer at all. As a result, the agent only learns from useful demonstrations even if they are diluted with useless experience. Our experiments in Chain (Section \ref{section_results_chain}) show that SILfD is unique among the LfD algorithms to possess this property.

\textbf{Automated scheduling.} Another challenge for the LfD algorithms is the setting of suboptimal expert, i.e. when its experience is consistent and useful but can be improved upon. In this setting, biasing the agent's goals towards expert imitation may prevent the agent from outperforming the expert. As \cite{jing2020trpofd} notice, efficiently dealing with this challenge requires to anneal the effect of demonstrations throughout the training, so that the expert behaviour could be consistently replicated without limiting further exploration. However, applying this technique to their or other existing LfD algorithms requires hand-crafted schedules and additional hyperparameter tuning. In contrast, SILfD automatically adjusts the effect of the demonstrations throughout the training depending on their current usefulness. Specifically, as the agent improves and the value function estimates increase, the contribution of demonstrations to the loss function (\ref{eq_our_policy}) decreases, whereas the agent's own experience becomes more useful. As a result, SILfD starts ignoring obsolete demonstrations once the agent performs on par with the expert. We confirm the ability of SILfD to consistently outperform the suboptimal expert in our experiments in DMC and report the schedules of the sampling probability of demonstrations discovered by SILfD in Section \ref{section_experiments_scheduling}.

\subsection{Demonstrations without rewards}

A drawback of SILfD is the assumption that the rewards are observed in demonstrations. This limits applicability of SILfD in the most interesting cases where demonstrations are collected by a human expert, possibly optimizing a different reward or acting in a different environment \cite{ziebart2008maximum,chentanez2018physics,scheller2020sample,pearce2021counter}. To mitigate this drawback, we modify SILfD when the rewards are unavailable in demonstrations. Instead of directly filling the replay buffer with demonstrations, we propose to first train BC (or other IL / offline RL algorithm) to mimic demonstrations and then initialize the buffer of SIL with its experience (Fig. \ref{fig:scheme_b}). We denote this modification as SILfBC. Additionally, we explore a variation of SILfBC where the parameters of the Actor of SIL are initialized via pretrained BC instead of the replay buffer, which we denote as BCSIL.

\begin{figure}[t]
\centering
\includegraphics[width=\linewidth]{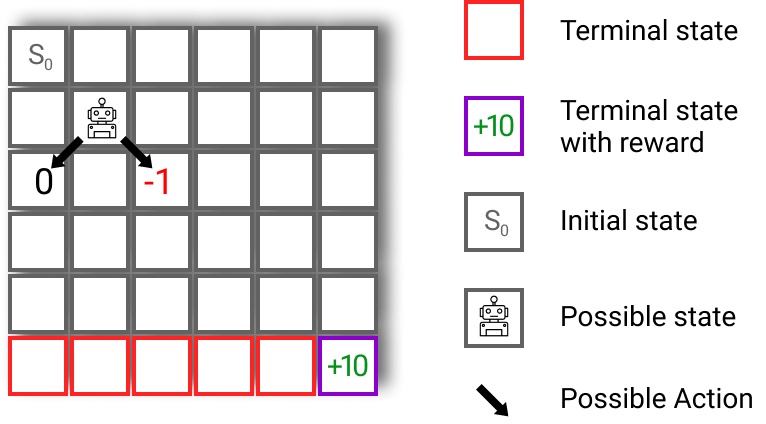}
\caption{Illustration of Chain environment}
\label{fig:chain_env}
\end{figure}

\subsection{Limitations of SILfD}

We highlight two of the possible problems with our approach: value overestimation and overdependence on the reward in demonstrations.

Overestimation of the lower-bound value by critic can occur for several reasons. First, stochasticity of policy or environment dynamics can cause high variance of return distribution in a given state, making critic estimate the highest rather than the expected return. This can be partially mitigated by using generalized SIL with n-step update \cite{tang2020self}. Second, since value approximations for all states are conditioned on the same vector of parameters $\phi$, for arbitrary states $s_1$ and $s_2$, updating $\phi$ to increase $V_\phi(s_1)$ can increase $V_\phi(s_2)$ as well, even if the latter is already tight. Third, naive initialization of $\phi$ can cause overestimation of the lower bounds in some states before the training even begins. While value overestimation is not specific to SILfD, it is partially alleviated in the original SIL by alternating with on-policy updates that can decrease overestimated values. In contrast, the issue can be exaggerated in SILfD: if demonstrations contain states that the agent does not reach, the value estimates of these states may rise uncontrollably.

A distinctive feature of SILfD is prioritization of demonstrations with high returns. While this feature makes SIL robust to suboptimal demonstrations, it can also backfire if demonstrations contain useful behaviour that does not reach any reward. For example, consider the task of stacking three cubes by a robot manipulator. If a demonstration of successfully stacking three cubes is provided, it will be used by SILfD to recover the demonstrated behaviour. However, if a demonstration of only stacking two cubes is provided and no reward is achieved, it will be deemed useless and ignored by SILfD. This can be mitigated by using generalized SIL in which the usefulness of demonstrations is not static, i.e. a transition can have a high priority due to leading to a state with a high value, even if no reward is observed.

\section{Experimental Procedure}

\subsection{Environments}

We have designed the experiments with the following desiderata in mind. First, SILfD should be tested in both discrete and continuous environments. To this end, we have chosen Pommerman and Chain as discrete environments and four DeepMind Control Suite tasks as continuous environments. Second, environments should have sparse rewards since in this setting demonstrations are the most helpful. While rewards in Chain and Pommerman are already sparse, we have additionally sparsified the rewards in DMC. Third, demonstrations should be imperfect to test both the robustness of algorithms and their ability to surpass the expert. To this end, we vary the proportion of suboptimal demonstrations mixed with one optimal demonstration in Chain and use a suboptimal expert in DMC. While the expert always wins in Pommerman, procedural map generation requires the agent to generalize beyond copying the expert.

\textbf{Chain} \cite{strens2000bayesian} is a simple but popular exploration benchmark \cite{osband2016deep,osband2018randomized}. The environment represents a square grid where the agent starts at the upper-left corner, its goal is to reach the bottom-right corner, and its actions are to move diagonally either to the lower-left cell or to the lower right cell. In our modification, the agent is penalized for moving right but receives a significant positive reward that exceeds any achievable sum of penalties by 100 when reaching the bottom-right corner. Additionally, if the agent is located on the left edge of the grid and steps left, it simply moves one cell down and does not receive a penalty. Without specific exploration strategies or relying on demonstrations, standard RL algorithms cannot find the positive reward and converge to the policy that avoids penalties by always moving left. We record one optimal demonstration where the expert only moves right and dilute it with $n \in [0, 99]$ adversarial demonstrations where the expert only moves left. We fix the size of the map at 40x40.

\begin{figure*}[t]
\centering
\includegraphics[width=\linewidth]{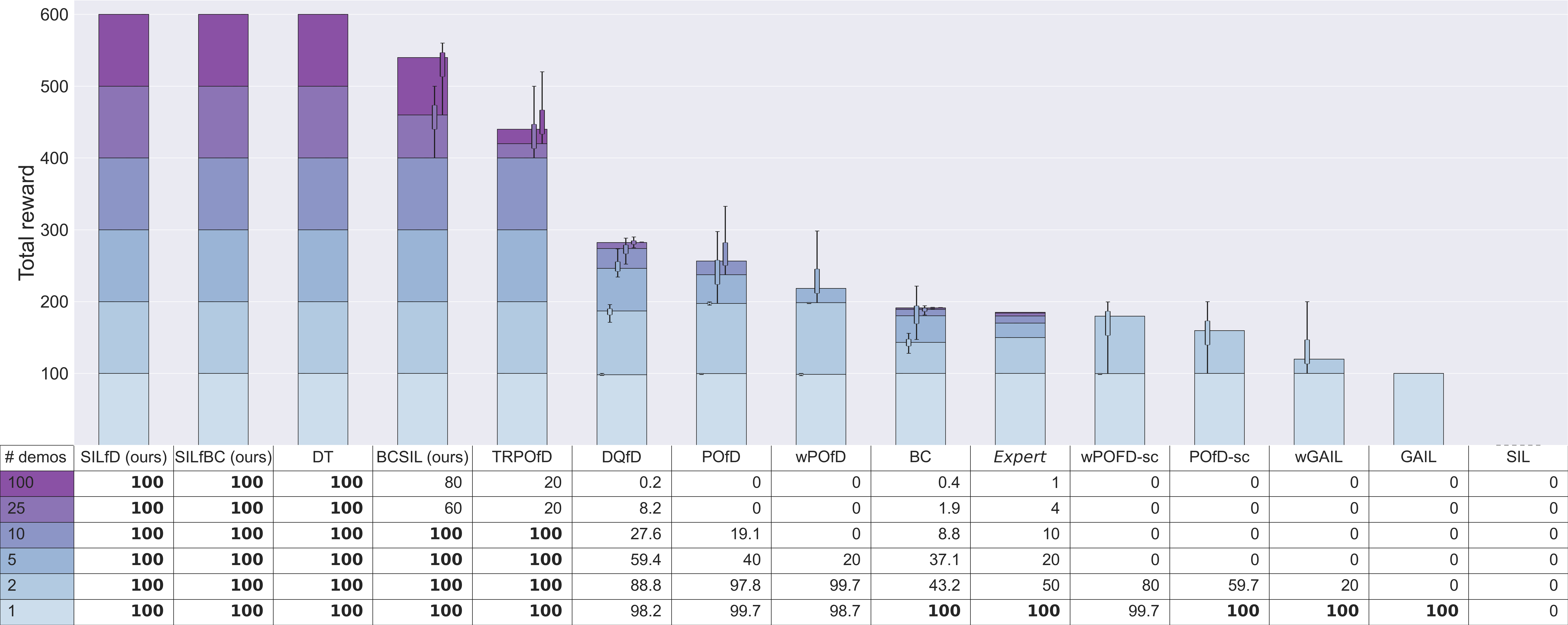}
\caption{Results in Chain. We evaluate six experimental settings where one optimal demonstration is mixed with a different number of adversarial demonstrations. The maximum reward in each setting equals 100. In the table, the rows represent the settings and the columns represent the algorithms. On the plot, Y-axis measures the total score over the settings, each represented by a bar of different color. Each experiment, i.e. an algorithm in a setting, is repeated for five seeds. Performance in each experiment is evaluated as the average over the seeds and over 100 episodes within each seed. The min-max spread of the performance within the seeds is reported for each experiment as a confidence interval, centered vertically at the average over the seeds and ordered horizontally in the increasing number of demonstrations. The algorithms are ordered in the decreasing summarized performance. The perfect scores are highlighted with bold font.}
\label{fig:results_chain}
\end{figure*}

\textbf{DeepMind Control Suite} \cite{tassa2018deepmind} is a set of popular benchmarks with continuous control. For our experiments, we select Cartpole Swing Up, which is a classic task where the agent needs to balance an unactuated pole by moving a cart, and three locomotive tasks: Cheetah Run, Walker Run and Hopper Hop, where the agent is supposed to control and to move forward a specific robot. In order to make exploration more challenging, we sparsify the rewards in all environments, the details of which are reported in the Appendix. We collect 25 suboptimal demonstrations with similar score in each environment. The details about the demonstrations are also reported in the Appendix.


\textbf{Pommerman} \cite{resnick2018pommerman} is a challenging multi-agent environment with discrete control and high-dimensional observations. We adapt the single-agent regime proposed in \cite{barde2020adversarial} where the agent needs to defeat a single random opponent. Similarly to \cite{barde2020adversarial}, we use the champion solution of FFA 2018 competition \cite{zhou2018hybrid} to gather 300 demonstrations. However, we do not modify the original environment used in the competition, including its procedural generation. This amplifies the difficulty for the algorithms as they have to generalize to unseen maps. Furthermore, the only reward signal the agent receives is +1 for defeating the opponent or -1 for being eliminated.

\subsection{Algorithms}

Each algorithm is tuned in each environment for 100 runs, each run repeated 2-3 times. The tuning procedure and the selected hyperparameters are reported in the Appendix.

\textbf{SILfD, SILfBC, BCSIL.} Our algorithms are implemented according to Section \ref{section_our}. In SILfD, the replay buffer is initialized with demonstrations. In SILfBC, the buffer is instead initialized with experience generated by pretrained BC. In BCSIL, a variation of SILfBC, instead of modifying the buffer, the weights of SIL are initialized as the weights of pretrained BC. Online updates of SIL are based on PPO.

\textbf{SIL.} Since vanilla SIL \cite{oh2018self} does not leverage demonstrations, comparing it with our algorithms highlights the benefits of demonstrations in sparse environments.

\textbf{BC.} This is a classic offline RL approach based on supervised learning that predicts a demonstrated action in a given state \cite{pomerleau1991efficient}. Since BC ignores rewards, it is unlikely to outperform the expert. 


\textbf{GAIL and POfD.} GAIL \cite{ho2016generative} is a modern IL algorithm that jointly trains a generator (policy) and a discriminator (reward predictor). POfD \cite{kang2018policy} is an extension of GAIL to the LfD setting that trains an Actor-Critic algorithm on a mixture of environmental and imitation rewards: $r = (1 - \lambda_1) r_{env} + \lambda_1 r_{im}, \lambda_1 \in [0, 1]$, where the imitation reward is predicted by a GAIL-like discriminator. As an Actor-Critic algorithm we use PPO. Because annealing the effect of suboptimal demonstrations may improve the performance, we also evaluate a version of POfD marked as \textit{POfD-sc} with a linear scheduling of $\lambda$.

\textbf{wGAIL and wPOfD.} wGAIL \cite{wang2021wgail} is a weighted modification of GAIL and the state-of-the-art in IL from suboptimal demonstrations. We additionally propose and evaluate its straightforward extension to the LfD setting, denoted as wPOfD, where the same mixture of rewards as in POfD is optimized, but the imitation reward is predicted by a wGAIL-like discriminator.

\textbf{DQfD and DDPGfD.} Similarly to SILfD, DQfD \cite{hester2018deep} and DDPGfD \cite{vecerik2017leveraging} store demonstrations in the replay buffer but apply several additional heuristics. Our implementations of DQN and DDPG are based on RLlib framework \cite{liang2018rllib}.

\textbf{TRPOfD.} The method proposed in \cite{jing2020trpofd} that we denote as TRPOfD is the state-of-the-art in the RL from suboptimal demonstrations. The algorithm imposes a hard constraint on the divergence from the expert policy that is relaxed overtime. Our implementation is based on a public implementation of TRPO.\footnote{\href{https://github.com/ikostrikov/pytorch-trpo}{github.com/ikostrikov/pytorch-trpo}}

\textbf{Decision Transformer.} Decision Transformer (DT) is a recent application of transformers to offline RL based on a framework that casts RL as conditional sequence modelling \cite{chen2021decision}. We use the authors' implementation.\footnote{\href{https://github.com/kzl/decision-transformer}{github.com/kzl/decision-transformer}}

\begin{table*}
\caption{Results in DMC and Pommerman. The rows represent the algorithms and the columns represent the environments. The format is $mean \pm std$, where mean and standard deviation are taken over five random seeds (DMC) or three random seeds (Pommerman), and the performance in each seed is the average over 100 episodes. The algorithms are ordered in the decreasing performance in Walker. The best score and the scores within the standard deviation of the best score are highlighted with bold font.}
\label{table:results_dmc_pom}
\vskip 0.15in
\begin{center}
\begin{small}
\begin{sc}
\begin{tabular}{lllllll}
\toprule
             & Type   & Cheetah                & Walker                 & Hopper                & Cartpole               & Pommerman              \\
\midrule
SILfD (ours) &  LfD  & 392.6 $\pm$ 38.75          & \textbf{502.32 $\pm$ 12.8} & \textbf{36.28 $\pm$ 8.22} & 674.48 $\pm$ 42.2         & 0.968 $\pm$ 0.003          \\
BCSIL (ours)  & LfD  & \textbf{441.8 $\pm$ 44.66} & \textbf{501.56 $\pm$ 8.49} & 18.1 $\pm$ 1.23           & 574.7 $\pm$ 288.7          & 0.326 $\pm$ 0.94           \\
SILfBC (ours) & LfD  & 316.56 $\pm$ 94.25         & \textbf{497.22 $\pm$ 6.23} & 21.41 $\pm$ 2.42          & \textbf{765.74 $\pm$ 23.2} & \textbf{0.976 $\pm$ 0.002} \\
TRPOfD       &  LfD  & 214.28 $\pm$ 149.9         & 481.1 $\pm$ 31.09          & 18.1 $\pm$ 22.21          & 0.0 $\pm$ 0.0              & 0.8 $\pm$ 0.22                       \\
POfD-sc     &   LfD  & \textbf{410.02 $\pm$ 49.05} & 373.58 $\pm$ 56.7         & 4.62 $\pm$ 2.23           & 1.8 $\pm$ 3.4              & -0.96 $\pm$ 0.03           \\
POfD        &   LfD  & 370.78 $\pm$ 87.38         & 324.8 $\pm$ 13.63          & 13.77 $\pm$ 1.45          & 18.86 $\pm$ 35.22          & -0.83 $\pm$ 0.12           \\
wPOfD-sc    &   LfD  & 41.32 $\pm$ 19.37          & 313.78 $\pm$ 8.72          & 6.48 $\pm$ 1.99           & 0.03 $\pm$ 0.07            & -0.97 $\pm$ 0.02           \\
BC         & Offline & 5.92 $\pm$ 1.18            & 310.18 $\pm$ 7.76          & 4.32 $\pm$ 1.39           & 2.12 $\pm$ 1.51            & -0.27 $\pm$ 0.09           \\
DT         & Offline & 14.0 $\pm$ 1.59            & 289.6 $\pm$ 4.2            & 8.21 $\pm$ 0.41           & 0.44 $\pm$ 0.61            & -0.96 $\pm$ 0.04           \\
\textit{Expert} & -  & 10.66 $\pm$ 11.05          & 287.96 $\pm$ 42.5          & 0.37 $\pm$ 0.39 & 63.07 $\pm$ 9.12           & 1.0 $\pm$ 0.0              \\
wPOfD       &   LfD  & 5.03 $\pm$ 1.3             & 282.66 $\pm$ 1.74          & 6.24 $\pm$ 3.13           & 6.22 $\pm$ 11.63           & -0.55 $\pm$ 0.36           \\
wGAIL       &   IL   & 5.9 $\pm$ 0.75             & 279.68 $\pm$ 8.62          & 5.28 $\pm$ 0.91           & 0.23 $\pm$ 0.47            & -0.999 $\pm$ 0.001         \\
DQfD / DDPGfD & LfD  & 94.33 $\pm$ 17.67          & 235.8 $\pm$ 132.5        & 1.65 $\pm$ 1.12           & 56.21 $\pm$ 17.81          & -0.88 $\pm$ 0.02           \\
GAIL    &       IL   & 5.79 $\pm$ 0.63            & 93.1 $\pm$ 9.26            & 6.68 $\pm$ 0.88           & 0.0 $\pm$ 0.0              & -1.0 $\pm$ 0.0             \\
SIL     &       RL   & 0.0 $\pm$ 0.0              & 70.78 $\pm$ 21.27          & 0.04 $\pm$ 0.01           & 0.0 $\pm$ 0.0              & -1.0 $\pm$ 0.0     \\
\bottomrule
\end{tabular}
\end{sc}
\end{small}
\end{center}
\vskip -0.1in
\end{table*}

\section{Experimental results}\label{section_experiments}

In this section, we first compare our methods with modern LfD, IL, and offline RL algorithms in the settings of noisy and suboptimal demonstrations, and then further explore the properties of SILfD.


\subsection{Comparison to existing algorithms}\label{section_experiments_sota}



\subsubsection{Chain.}\label{section_results_chain} 

The results are presented in Figure \ref{fig:results_chain}. In these experiments, we dilute one optimal expert trajectory with adversary demonstrations.

We find that only our algorithms and DT perform consistently in all settings. For SILfD and SILfBC, this points towards their robustness to the useful demonstrations being diluted. Furthermore, for SILfBC the performance of BC is crucial. While overall BC approximates the expert well, in the hardest setting its score is close to 0, making its experience almost always useless. However, even a rare successful episode is sufficient for SILfBC. BCSIL also leverages pretrained BC successfully but is not as stable in harder settings. Regarding DT, its flawless performance is expected since it is designed to distinguish demonstrated behaviours that lead to different returns.

The performance of the existing LfD algorithms falls off in harder settings. TRPOfD performs the best among them as it solves the easier settings consistently. However, the two hardest settings are only occasionally solved: the constraint on the divergence from the expert policy may initially force the agent to prioritize going left, which may persist even as the constraint relaxes due to the lack of the experience of going right. POfD and DQfD outperform the expert in most settings but follow the same trend of performing worse with the increased number of adversarial demonstrations. The weighted version wPOfD fails to outperform even the original POfD. Contrary to our expectations, scheduling the reward mixture coefficient negatively affects POfD and wPOfD, which might be due to the objectives becoming inconsistent. These results support our discussion in Section \ref{section_our_properties} about the existing algorithms having difficulties with discerning the usefulness of demonstrations.

Finally, the IL algorithms GAIL and wGAIL fail to even replicate the expert performance, which might be due to the adversarial training being unstable in the multi-modal regimes. Regarding wGAIL, while the algorithm is designed for suboptimal demonstrations, it relies on the optimal part of demonstrations being more consistent, whereas in our Chain experiments both optimal and adversarial demonstrations are deterministic and hence equally consistent.

\subsubsection{DMC.}\label{section_results_dmc}

The results are presented in Table \ref{table:results_dmc_pom}. Overall, our methods achieve the best results across the environments. We hypothesize that the automated scheduling is crucial for their performance. LfD algorithms without scheduling have to compromise between exploiting demonstrations early in the training and limiting their final performance. On the other hand, simple hand-crafted schedules can still be suboptimal even with tuned hyperparameters. Likewise, IL algorithms and BC generally fail to outperform the expert, while DT only succeeds occasionally. Finally, SIL shows worst results as it struggles to find any meaningful behaviour due to the sparsity of the reward.


\subsubsection{Pommerman}\label{section_results_pommerman}

The results are presented in Table \ref{table:results_dmc_pom}. We find that only our algorithms SILfD and SILfBC successfully recover expert performance and consistently win in procedurally generated maps. BCSIL, a variation of SILfBC with pretrained weights, may suffer from the catastrophic forgetting as it only successfully trains in two of three seeds. Offline algorithms BC and DT manage to win a few matches in unseen maps, the former being sufficient for SILfBC to explore the winning solution. The only other algorithm to perform well is TRPOfD. However, it struggles to stabilize its final performance, which is likely due to the bias towards the expert behaviour. Finally, all other LfD and Imitation Learning algorithms fail to find the optimal solution as the environment requires generalization to unseen layouts of the map.

\subsection{Additional experiments}\label{section_results_additional}

\begin{figure}[t]
    \centering
    \begin{subfigure}{0.95\linewidth}
        \centering
        \includegraphics[width=\linewidth]{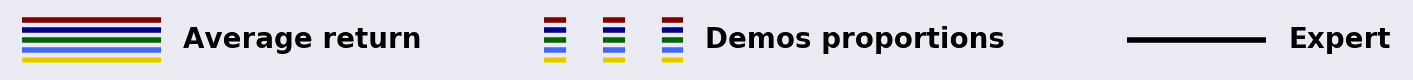}
        \label{fig:legend_schedule}
    \end{subfigure}
    \begin{subfigure}{.49\linewidth}
        \centering
        \includegraphics[width=\linewidth]{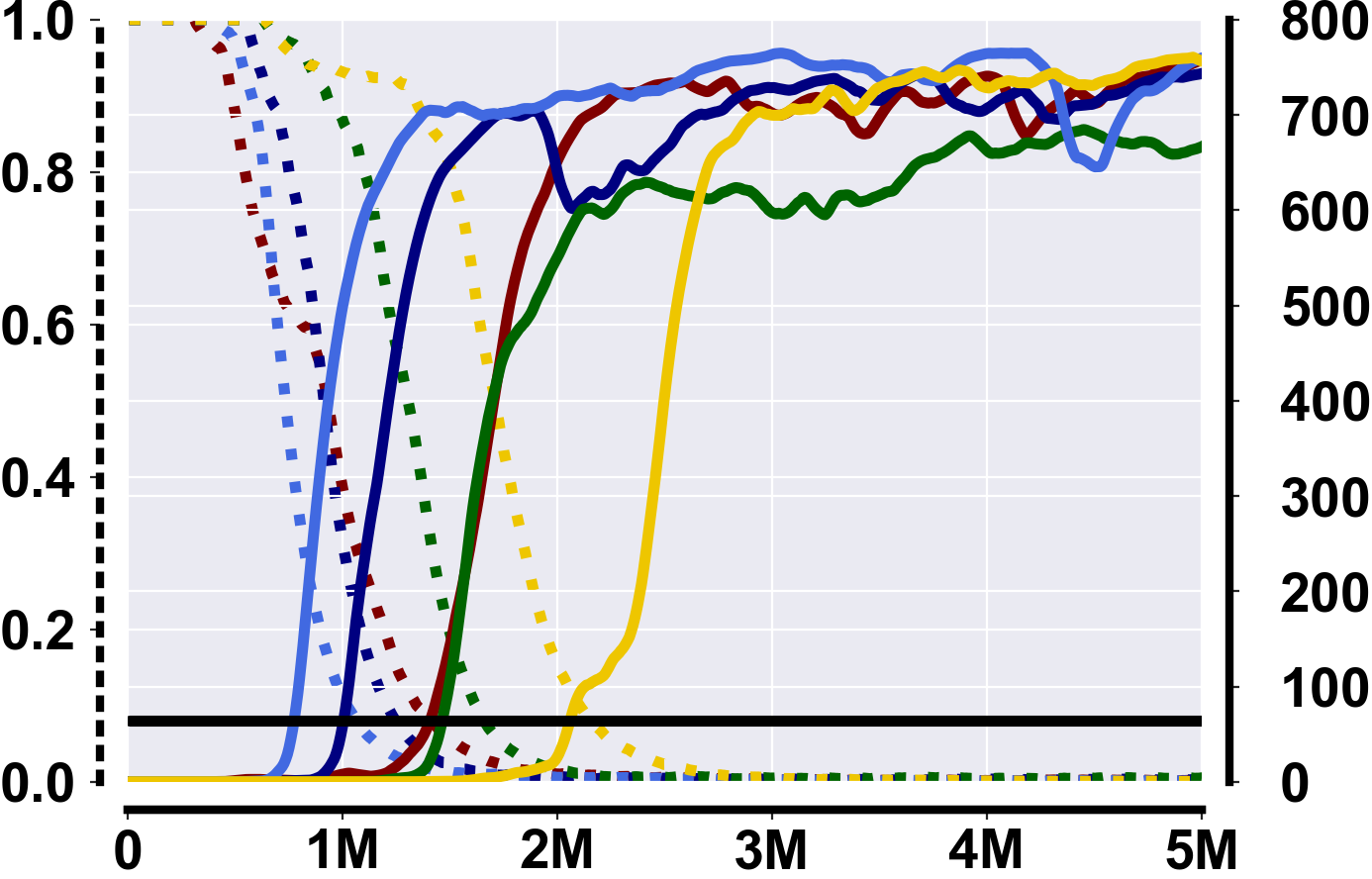}
        \caption{Cartpole}
        \label{fig:cartpole_schedule}
    \end{subfigure}
    \begin{subfigure}{.49\linewidth}
        \centering
        \includegraphics[width=\linewidth]{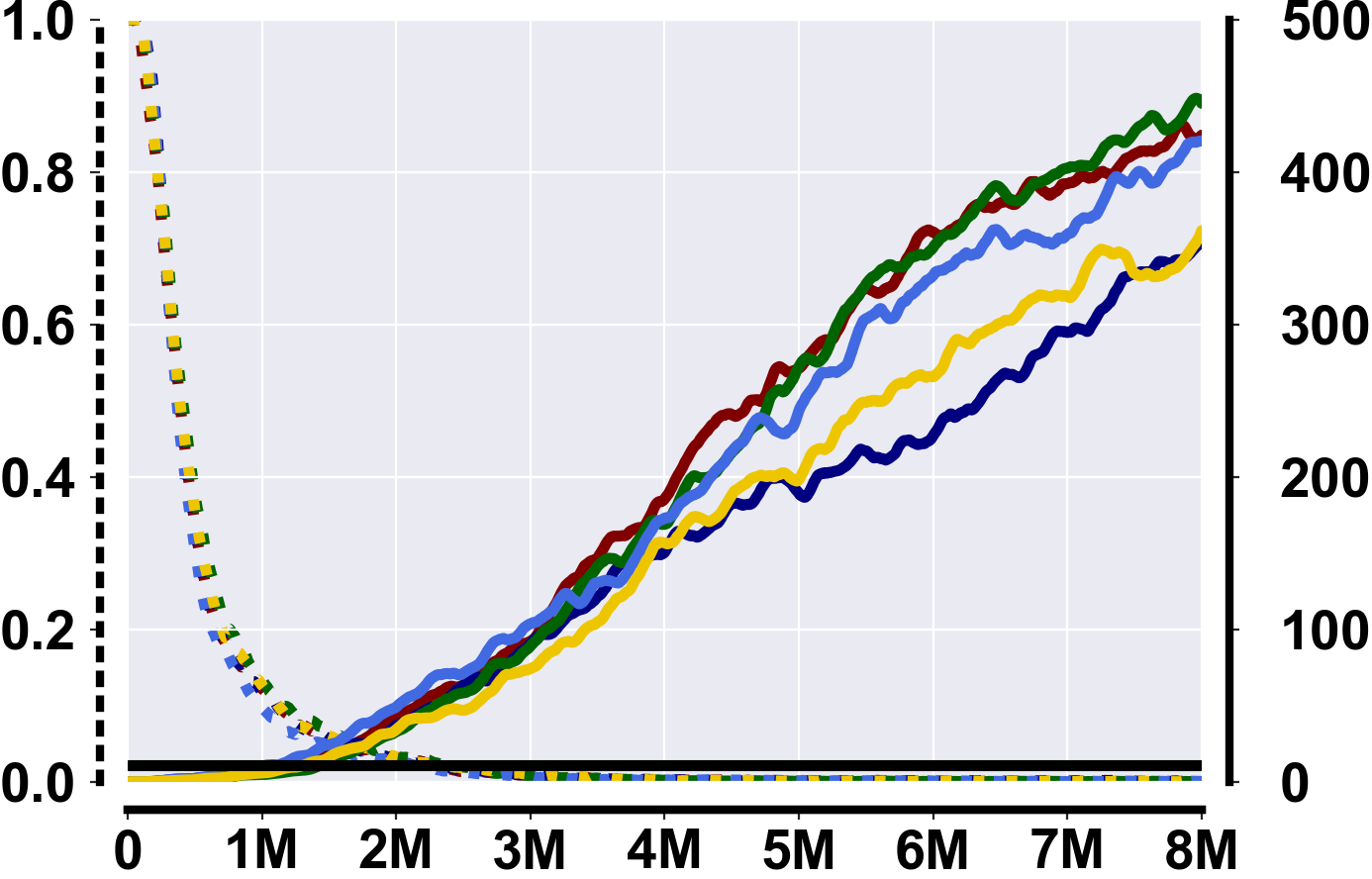}
        \caption{Cheetah}
        \label{fig:cheetah_schedule}
    \end{subfigure}
    \begin{subfigure}{.49\linewidth}
        \centering
        \includegraphics[width=\linewidth]{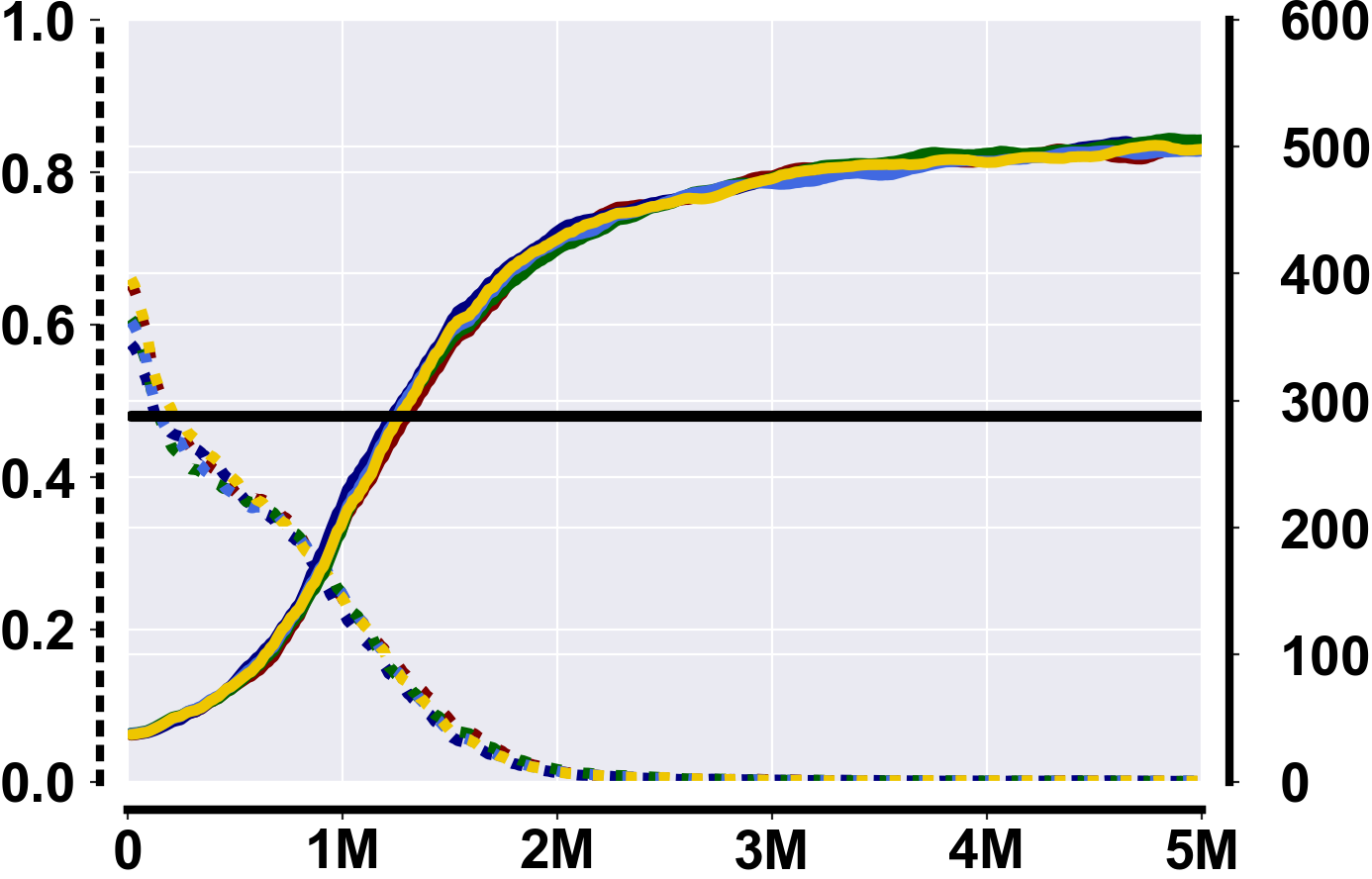}
        \caption{Walker}
        \label{fig:walker_schedule}
    \end{subfigure}
    \begin{subfigure}{.49\linewidth}
        \centering
        \includegraphics[width=\linewidth]{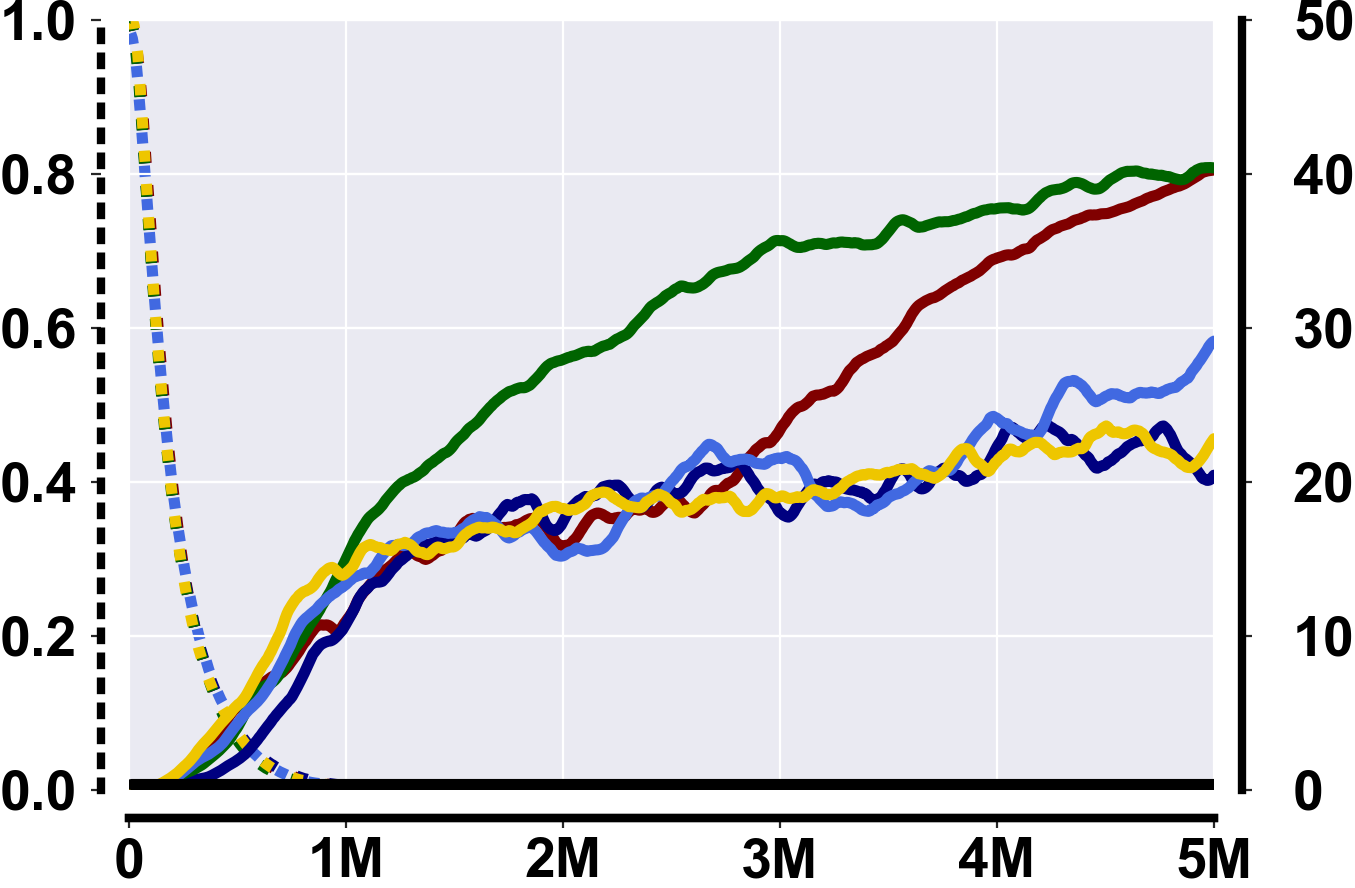}
        \caption{Hopper}
        \label{fig:hopper_schedule}
    \end{subfigure}
    \caption{Automated scheduling in SILfD. Curves of the same color correspond to the same seed. Left Y-Axis corresponds to the dashed curves and measures the proportion of the demonstrated data in the batches sampled during SIL updates. Right Y-axis corresponds to the solid curves and measures the average return over 100 episodes. Solid black line indicates the expert performance.}
    \label{fig:scheduling_results}
\end{figure}

\begin{figure}[t]
    \centering
    \begin{subfigure}{\linewidth}
        \centering
        \includegraphics[width=\linewidth]{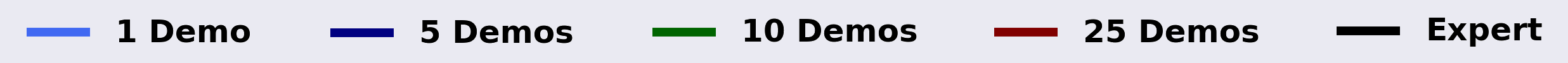}
        \label{fig:legend_demo_num}
    \end{subfigure}
    \begin{subfigure}{.448\linewidth}
        \centering
        \includegraphics[width=\linewidth]{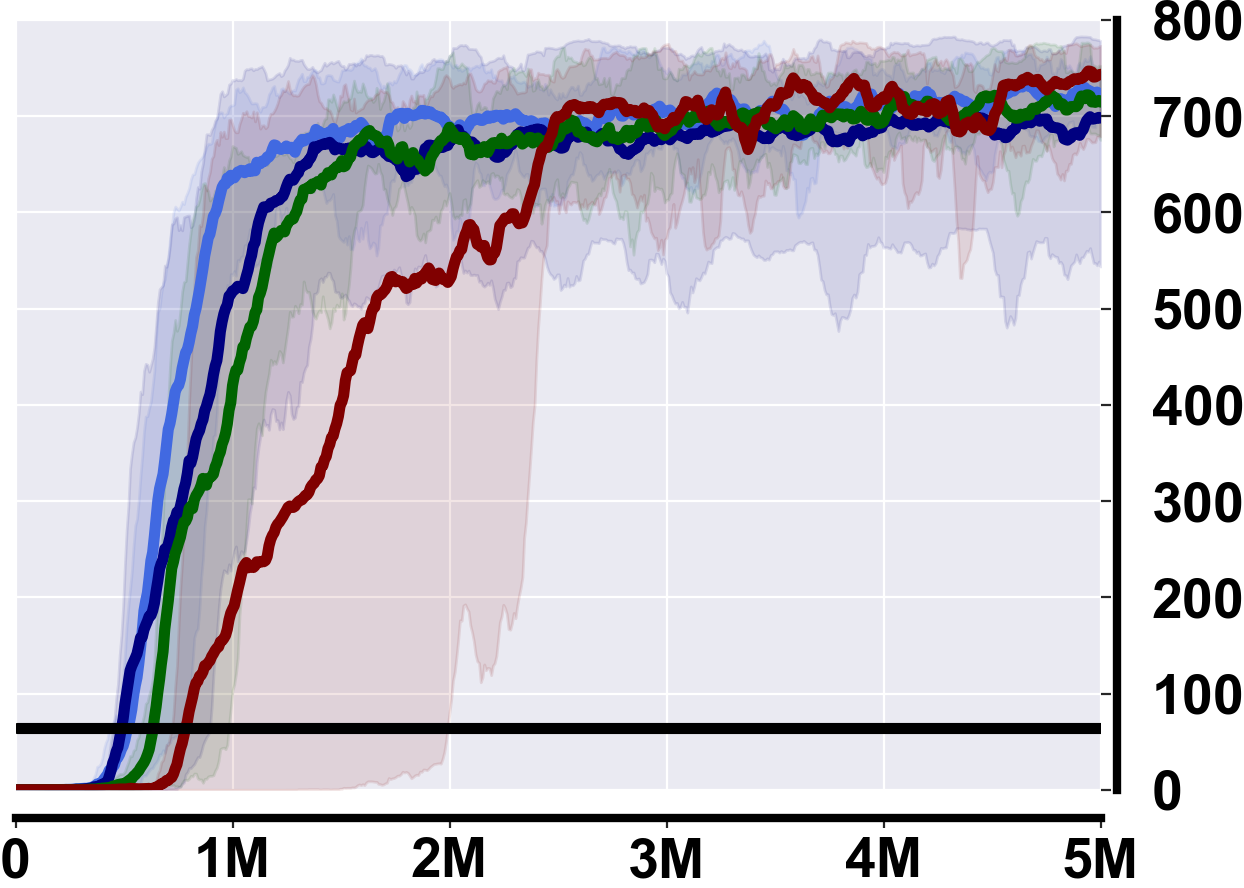}
        \caption{Cartpole}
        \label{fig:cartpole_demo_num}
    \end{subfigure}
    \begin{subfigure}{.448\linewidth}
        \centering
        \includegraphics[width=\linewidth]{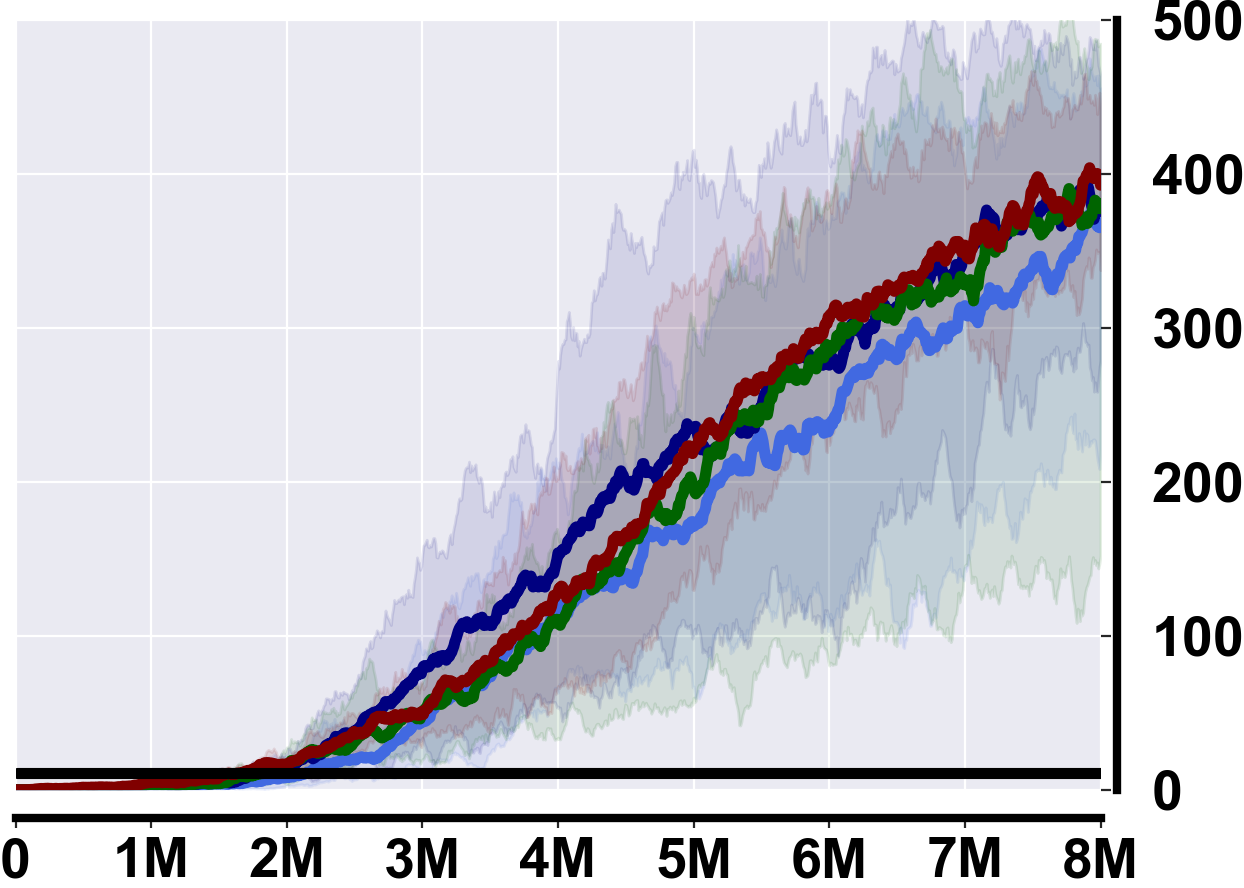}
        \caption{Cheetah}
        \label{fig:cheetah_demo_num}
    \end{subfigure}
    \begin{subfigure}{.446\linewidth}
        \centering
        \includegraphics[width=\linewidth]{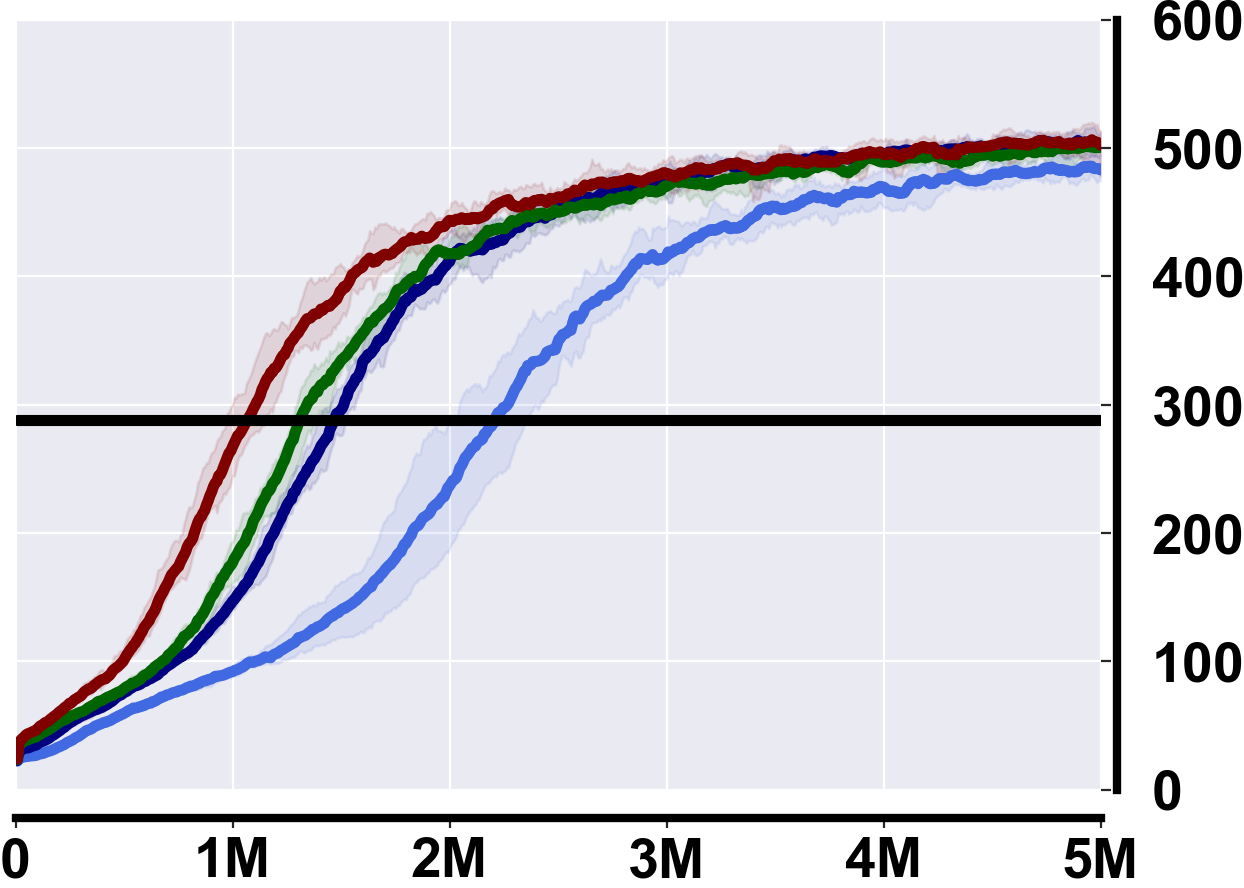}
        \caption{Walker}
        \label{fig:walker_demo_num}
    \end{subfigure}
    \begin{subfigure}{.446\linewidth}
        \centering
        \includegraphics[width=\linewidth]{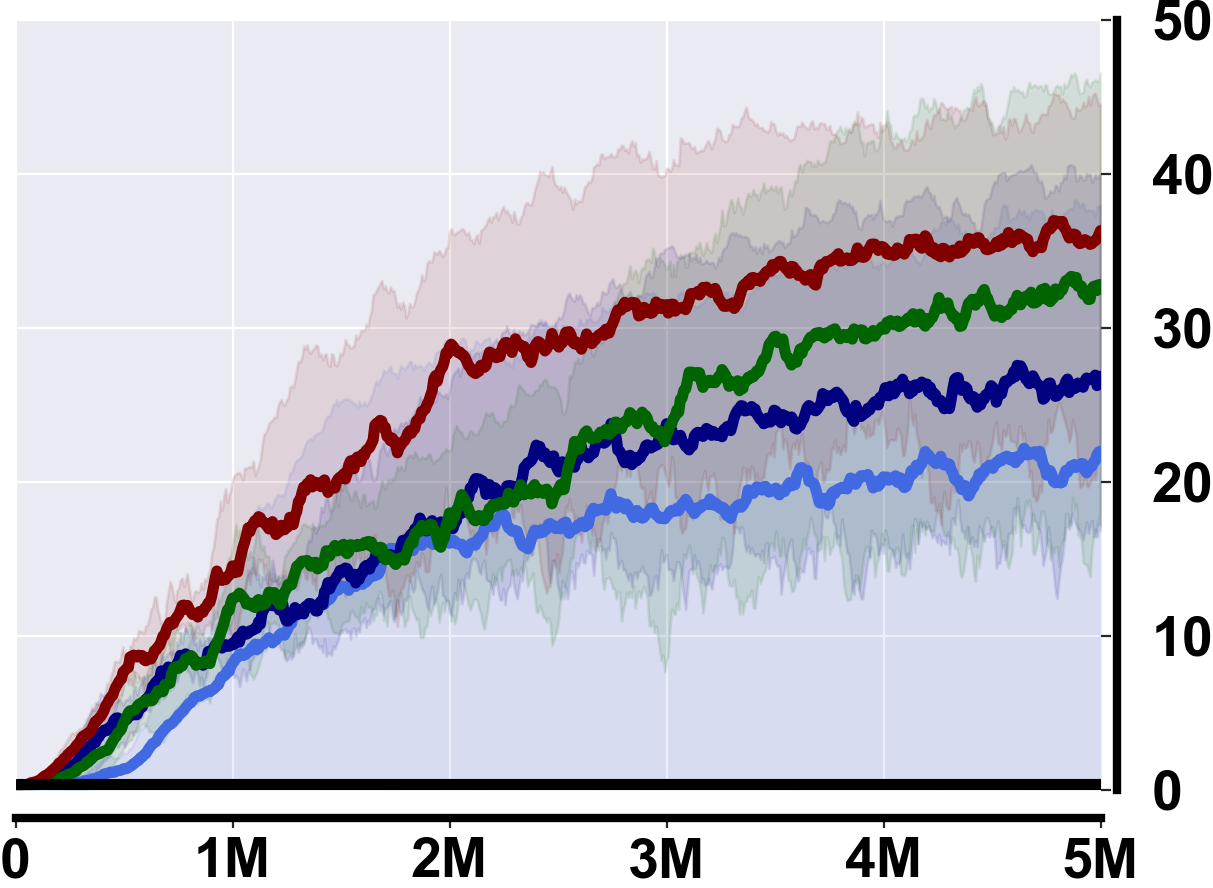}
        \caption{Hopper}
        \label{fig:hopper_demo_num}
    \end{subfigure}
    \caption{SILfD scores for varied number of demonstrations. Y-Axis measures the average return during the last 100 episodes. For each specific number of demonstrations, we run 5 trials. The solid curves corresponds to mean score among trials. Semitransparent areas correspond to min-max intervals. Solid black line indicates the expert performance.}
    \label{fig:demos_number}
\end{figure}

\subsubsection{Automated scheduling in SILfD}\label{section_experiments_scheduling}

In this experiment we test the ability of SILfD to automatically schedule the influence of the demonstrations on the learning process discussed in Section \ref{section_our_properties}. We train SILfD and track the proportion of the expert transitions in the batches sampled from the buffer during SIL updates. The proportion being equal to 1 corresponds to the batch consisting solely of the demonstrated data and vice versa. We run this experiment in each DMC environment for 5 seeds. The results are presented in Figure \ref{fig:scheduling_results}. We find that during the beginning of the training when the agent fails to find any meaningful experience in the environment, the proportion of the demonstrations is close to 1. As the agent improves, the proportion value begins to decrease. This connection of the proportion and the agent's performance is the most visible in Cartpole (\ref{fig:cartpole_schedule}). In all experiments we observe that as the agent starts to outperform the expert, the proportion rapidly declines and then stays at zero. This confirms our intuition that SILfD automatically anneals the influence of the demonstrations throughout the training. It is also interesting to note that the discovered schedule depends on the random seed, which is difficult to hand-code.

\subsubsection{Varying number of demonstrations}\label{section_experiments_ndemos}

We now validate the ability of SILfD to learn from limited experience. For each DMC environment, we select the subsets of demonstrations containing 1, 5, 10, and all 25 episodes. In each subset, the trajectories are chosen by the proximity to the average expert performance. We train SILfD five times on each subset. The results are reported in Figure \ref{fig:demos_number}. We observe that SILfD is capable of outperforming the expert by a margin regardless of the number of the demonstrations. Furthermore, in all environments except Hopper (\ref{fig:hopper_demo_num}), varying the number of the demonstrations does not influence the final performance of the agent. This experiment highlights the ability of SILfD to benefit from the demonstrations even in the settings without access to large data sets of expert trajectories.

\section{Conclusion}

In this paper, we present SILfD, a novel algorithm that incorporates demonstrations into Self-Imitation Learning. We formulate and experimentally verify its crucial properties that are unique among the LfD algorithms, namely the ability to discern the most useful demonstrated behaviour and the automated scheduling of the influence of demonstrations. We show that SILfD reliably surpasses the expert and achieves state-of-the-art performance in learning from suboptimal or noisy demonstrations. Additionally, we propose SILfBC, a modification of SILfD that relaxes the requirement to observe rewards in demonstrations. We find it to be competitive with existing state-of-the-art approaches, making it a viable alternative to the original algorithm.

\bibliography{main}
\bibliographystyle{aaai_style/aaai}

\setcounter{section}{0}
\begin{appendix}
\appendix
\title{Appendix}
\maketitle

\section{Environments}


Here we report technical details of the environments.

\subsection{Chain}

In Chain Environment, observation space is represented by a square grid. Agent's state in the environment is defined by its position on this grid and consists of two components: horizontal and vertical. Each component is normalized to be in range from $-1$ to $1$. Thus, the initial state in the environment is $[-1, -1]$, and for all the terminal states the second component is $1$. The goal of the agent is to reach state $[1, 1]$, i. e. the bottom right corner of the grid. Action space is discrete and specifies two options: going left ($0$) and going right ($1$). The total reward for the episode lies within the range $[-60, 100]$.

Each algorithm was trained for 1 million transitions in the environment, which constitutes up to 30 minutes of wall-clock time, depending on the algorithm. The exceptions are DQfD, which is more sample efficient and was trained for 200 thousands transitions, and BC and DT, which do not require interacting with the environment. For SILfBC, we collect 1000 BC demonstrations to fill the buffer of SIL.

\subsection{DeepMind Control Suite}

In the original Cartpole, a non-zero reward is given every time the pole is positioned vertically and the cart is located within a certain range. In our modification, the agent does not receive the reward if the velocities of the pole or the cart exceed a threshold. As a result, it becomes extremely difficult for the agent to find the learning signal randomly. In order to collect demonstrations, we handcraft a heuristic PID-controller that swings the cart back and forth until the pole reaches a vertical position. In the locomotive tasks, the reward is originally given each time the robot advances forward. In our modifications, the agent receives the reward proportional to its velocity, but only if the velocity exceeds a certain threshold. As an expert, we train a PPO agent to move with a target speed that is only slightly higher than the threshold. Since it is possible to move faster than such expert, demonstrations of its behaviour are suboptimal for the agent.

For SILfBC, we collect 25 BC demonstrations to fill the buffer of SIL in each DMC environment. Note that we filter these demonstrations and only select those with non-zero return.

\subsubsection{Cartpole}
The state space is defined by $5$ components: position of the cart, cosine of the pole, sine of the pole and cart's and pole's velocities. Action space is continuous and one-dimensional. Action's magnitude specifies the amount of force
applied to move a cart and the sign of the action specifies the direction of the force. In the original implementation of the environment the reward is given whenever the horizontal position of the cart is within a range of from $-0.25$ to $0.25$ and the pole's cosine is within a range from $0.995$ to  $1$. In our modification agent does not receive any reward whenever the pole's and cart's velocities exceed $0.25$ and $0.5$ respectively. To collect demonstrations in this environment, we handcrafted a heuristic PID-controller. It receives $13$ total reward per episode on average. However, we filtered demonstrations containing only those episodes where the total reward exceeded $40$, and thus the expert's performance in demonstrations is approximately $65$.

Each algorithm except BC and DT is trained for 5 million transitions in the environment.

\subsubsection{Walker}
In the Walker environment, the agent controls the robot that has two actuated legs. The action space consists of 6 components, where each component represents the force applied to a certain joint of a single leg. The state space consists of 24 components that represent positions, orientations and velocities of different parts of the robot. In the original implementation of the environment the reward is divided into two parts: the first part is the standing reward $r_s$ which increases towards 1 when the vertical position of the agent's torso gets closer to the value of 1.2; the second part is the moving reward $r_m$ which increases up to 1 as the agent's horizontal velocity gets closer and exceeds the threshold of 8. The final reward is calculated using the following equation: 

\begin{equation}
    r = r_s \frac{(5 r_m + 1)}{6}
\end{equation}

In order to collect demonstrations for experiments in Walker, we modify the moving part of the original reward. To train the agent run with a certain target speed, we set the reward to be 1 if the velocity of the agent lies inside the interval between 4 and 5 and we decrease the reward as it gets further from its boundaries. We train agent to maximize this reward using PPO algorithm.

To create sparse version of the environment, we modify the moving reward to be 0 if the agent's velocity is below the threshold of 4. In order to encourage the agent to run faster as it exceeds this threshold, we linearly increase the reward depending on agent's velocity.

Each algorithm except BC and DT is trained for 5 million transitions in the environment.

\subsubsection{Hopper}
In the Hopper environment, the agent controls the robot that has a single leg. The state and action spaces are represented by the vectors with 15 and 4 components respectively. The reward scheme is the same as in the Walker Environment, but the target speed for the expert lies in the interval between 1 and 2. In the sparse version of the environment the speed threshold is set to be 1.3.

Each algorithm except BC and DT is trained for 5 million transitions in the environment.

\subsubsection{Cheetah}
In the Cheetah environment the agent controls the robot with two legs: back and front. The state and action spaces are represented by the vectors with 17 and 6 components respectively. The reward scheme is similar to Hopper's and Walker's reward scheme, but the standing reward is absent. The target speed for the expert lies in the interval between 5 and 6, and the velocity threshold in sparse version of the environment is set to be 5.

Each algorithm except BC and DT is trained for 10 million transitions in the environment.

\subsection{Pommerman}

The observation space is composed of a 11x11 grid with 15 one-hot features, 2 feature maps, and an additional information vector. The one-hot feature represents an element on the map. Specifically, these features can represent  the current player, an ally, an enemy, a passage, a wall, wood, a bomb, flame,
fog, and a power-up. The feature maps contain integers indicating bomb blast strength and bomb life for each location. Finally, the additional information vector contains
the time-step, number of ammunition, whether the player can kick and blast strength for the current
player. The agent has six actions: do-nothing, up, down, left, right, and lay bomb.

We use Agent47Agent to gather expert trajectories, which is based on Monte-Carlo Tree Search.\footnote{\href{https://github.com/YichenGong/Agent47Agent/tree/master/pommerman}{github.com/YichenGong/Agent47Agent/tree/master/pommerman}} We only select winning trajectories from the expert. 

Each algorithm was trained for 10 million transitions in the environment. The exceptions are DQfD, which is more sample efficient and was trained for 1 million transitions, and BC and DT, which do not require interacting with the environment.

For SILfBC, we collect 400 BC demonstrations to fill the buffer of SIL. These are not filtered.

\section{Algorithms and Hyperparameters}

Below we describe how hyperparameters were tuned. The selected hyperparameters are reported in Tables \ref{tab:hyperparameters}, \ref{tab:hyperparameters2}.

In all environments and for each algorithm, we select hyperparameters using a random search procedure. Specifically, we select a 100 random hyperparameter configurations, train an algorithm with each configuration on 2 random seeds (Pommerman) or 3 random seeds (Chain, DMC), evaluate the trained algorithm for 100 episodes, and select the configuration with the highest average score over the seeds and the episodes. We then rerun each experiment 3 times (Pommerman) or 5 times (Chain, DMC) with the selected hyperparameters, the results of which are reported in the main text.

In Chain, we tune hyperparameters in the setting with one optimal and one adversary demonstrations provided (2 in total) and apply these hyperparameters in the other settings. We select hyperparameters for Pommerman and each environment in DMC independently.

Below we report the tuning ranges and the additional details for each algorithm.

All networks except DT have the following architectures: [in\_dim, 32, 32, out\_dim] in Chain, [in\_dim, 256, 256, out\_dim] in DMC, the architecture identical to \cite{barde2020adversarial} in Pommerman. All networks have ReLU activation functions between the layers and are trained with Adam optimizer \cite{kingma2015adam}.

\textbf{PPO.} PPO is not a separate baseline but is used as a basic algorithm in the online updates of SIL, SILfD, SILfBC and BCSIL, as well as in GAIL, POfD, POfD (sc) and their weighted versions. For each of these algorithms, we select the following PPO hyperparameters: learning rate from $\{1e-4, 2e-4, 5e-4, 1e-3\}$, batch size from $\{128, 256, 512\}$, epochs per update from $\{3, 10, 30\}$. We also employ Generalized Advantage Estimation (GAE) during training \cite{schulman2015high}.

\textbf{SIL, SILfD, SILfBC, BCSIL.} On top of the PPO hyperparameters reported above, for SILfD we select: epochs per SIL update from $\{5, 10, 20, 40\}$, SIL loss weight from $\{0.01, 0.1, 1, 10\}$, SIL batch size from $\{256, 512\}$. Online updates in SIL are based on PPO in all environments. We do not separately tune SIL, SILfBC and BCSIL and instead use the hyperparameters selected for SILfD and BC.

\textbf{BC.} For BC, we select learning rate from $\{5e-5, 1e-4, 2e-4, 5e-4, 1e-3, 2e-3\}$ and batch size from $\{64, 128, 256, 512\}$.

\textbf{GAIL, POfD, wGAIL, wPOfD.} On top of the PPO hyperparameters reported above, for POfD we select discriminator batch size from $\{128, 256, 512\}$, discriminator epochs per update from $\{3, 10\}$, discriminator learning rate from $\{5e-5, 1e-4, 2e-4, 5e-4\}$, and reward mixing coefficient $\lambda_1 \in \{0.01, 0.1, 0.5, 0.9, 0.99\}$. For GAIL, we use the hyperparameters selected for POfD but remove the environmental reward and $\lambda_1$. For POfD (sc), we also use the hyperparameters selected for POfD, but initially set $\lambda_1$ to 1 and gradually anneal it according to a linear schedule $\lambda_1^{new} = \max(0, \lambda_1 - 1 / (\text{num\_updates} * \lambda_1^{sc}))$. The inverse speed of annealing $\lambda_1^{sc}$ is additionally selected from $\{0.2, 0.5, 1, 2, 5\}$ with other hyperparameters fixed. For the weighted versions of the algorithms, the tuning procedure is identical, except that an additional hyperparameter $\beta$ is selected from $\{0, 1, 2\}$.

\textbf{DQfD, DDPGfD.} For DQfD and DDPGfD, we select learning rate from $\{1e-4, 2e-4, 5e-4, 1e-3\}$, batch size from $\{128, 256, 512\}$, n-step loss weight from $\{0.1, 1\}$, l2 regularization weight from $\{1e-5, 1e-4, 1e-3\}$, priority bonus of demonstrations from $\{0.1, 0.2, 0.5, 1\}$. We apply dueling \cite{wang2016dueling} and double \cite{van2016deep} modifications of DQN in DQfD. We train DQfD and DDPGfD 5 times as less as other algorithms to utilize their sample efficiency and prevent overfitting.  

\textbf{TRPOfD.} For TRPOfD, we select maximum allowable KL divergence from the previous policy aka size of the trust region $\delta$ from $\{1e-4, 2e-4, 5e-4, 1e-3, 2e-3, 5e-3, 1e-2\}$, initial maximum divergence from the expert policy $d_0$ from $\{1e-4, 1e-3, 1e-2\}$, exponential annealing factor $\epsilon$ from $\{1e-5, 1e-4, 2e-4, 5e-4, 1e-3, 2e-3\}$, the learning rate of the recovery objective from $\{0.01, 0.1, 1\}$. This algorithm processes whole batch at once during the update. The annealing of the divergence from the expert $d_k$ is according to the exponential rule: $d_{k+1} = d_k + d_k \epsilon$, where $k$ is the epoch number. Each update, the whole batch is processed at once  a single time.

The original algorithm employs MMD to measure the divergence from the expert, however, we instead use the KL divergence. In effect, we maximize the log-likelihood of the expert actions under the agent policy, which is similar to BC. There are two reasons for this change. First, MMD only for continuous environments, whereas our modification can also be applied in discrete environments. Second, in the preliminary experiments we found our modification to perform better even in continuous environments.

\textbf{DT.} For Decision Transformers, we use the hyperparameters reported in the original paper for D4RL (which contains analogues of DMC environments). We also simplify the architecture of DT in Chain by reducing the number of attention blocks.

During evaluation, DT requires to specify the desirable returns. In each environment, we select several desirable returns, evaluate trained DT with each of the selected returns, and report the best performance. In Chain, the selected returns are 10 and 100. In Pommerman, the selected returns are 0.5 and 1. In DMC, we select ten uniformly spaced values from 0 to some maximal threshold (0 is not included). Specifically, the maximal threshold equals 500 in Cheetah and Walker, 50 in Hopper, and 800 in Cartpole.

\begin{table*}[t]
\centering
\caption{Hyperparameters for all environments and algorithms (continued below)}
\begin{tabular}{lcccccc}
\hline
Hyperparameter                                               & \multicolumn{1}{l}{Chain} & \multicolumn{1}{l}{Pommerman} & \multicolumn{1}{l}{Walker} & \multicolumn{1}{l}{Hopper} & \multicolumn{1}{l}{Cheetah} & \multicolumn{1}{l}{Cartpole}  \\ \hline
                                      &     & & & & \\
\hspace{3mm}\textbf{General parameters}              &     & & & & \\
\hspace{3mm}Transitions between updates                & 1000  & 1000       & 1000   & 1000   & 1000                         & 1000     \\
\hspace{3mm}Number of workers                          & 1     & 8         & 8      & 8      & 8                            & 8        \\
\hspace{3mm}Discounting factor $\gamma$                                   & 0.99  & 0.99      & 0.99   & 0.99   & 0.99                         & 0.99     \\
                                      &     & & & & \\
\hspace{3mm}\textbf{PPO}              &     & & & & \\
\hspace{3mm}Entropy coefficient                        & 0.01  & 0.01      & 0      & 0      & 0                            & 0        \\
\hspace{3mm}Update clipping parameter                  & 0.2   & 0.2       & 0.2    & 0.2    & 0.2                          & 0.2      \\
\hspace{3mm}GAE $\lambda$                              & 0.95  & 0.95      & 0.95   & 0.95   & 0.95                         & 0.95     \\
                                      &     & & & & \\
\hspace{3mm}\textbf{SIL, SILfD, SILfBC, BCSIL}              &     & & & & \\
\hspace{3mm}Batch size                                 & 32    & 512       & 512    & 128    & 512                          & 256      \\
\hspace{3mm}Learning rate                              & 2e-4  & 1e-4      & 1e-4   & 2e-4   & 1e-4                         & 1e-4     \\
\hspace{3mm}Epochs per update                          & 3     & 3         & 30     & 10     & 10                           & 30       \\
\hspace{3mm}SIL value loss weight $\beta$              & 0.01  & 0.1       & 0.1    & 0.1    & 0.1                          & 0.1      \\
\hspace{3mm}Epochs per SIL update                      & 40    & 40        & 40     & 10     & 5                            & 20       \\
\hspace{3mm}SIL loss weight                            & 10    & 1         & 1      & 10     & 0.1                          & 0.1      \\
\hspace{3mm}SIL Batch size                             & 256   & 256       & 512    & 512    & 512                          & 256      \\
\hspace{3mm}Buffer size                                & 1e5   & 1e5       & 1e5    & 1e5    & 1e5                          & 1e5      \\
                                      &     & & & & \\
\hspace{3mm}\textbf{GAIL, POfD}              &     & & & & \\
\hspace{3mm}Batch size                                 & 32    & 256       & 256    & 512    & 512                          & 512      \\
\hspace{3mm}Learning rate                              & 2e-5  & 5e-4      & 1e-4   & 1e-4   & 1e-4                         & 2e-4     \\
\hspace{3mm}Epochs per update                          & 3     & 3         & 10     & 10     & 10                           & 30       \\
\hspace{3mm}Discriminator learning rate                & 1e-5  & 5e-5      & 2e-4   & 2e-4   & 2e-4                         & 5e-4     \\
\hspace{3mm}Discriminator batch size                   & 32    & 128       & 512    & 512    & 128                          & 128      \\
\hspace{3mm}Discriminator epochs per update            & 3     & 10        & 10     & 10     & 10                           & 10       \\
\hspace{3mm}Mixing coefficient $\lambda_1$ (POfD)      & 0.9   & 0.01      & 0.5    & 0.01   & 0.01                         & 0.01     \\
\hspace{3mm}Annealing coef $\lambda_1^{sc}$ (POfD sc)  & 5     & 1         & 0.5    & 0.2    & 0.2                          & 1        \\
                                      &     & & & & \\
\hspace{3mm}\textbf{wGAIL, wPOfD}              &     & & & & \\
\hspace{3mm}Batch size                                 & 32    & 256       & 512    & 128    & 512                          & 128      \\
\hspace{3mm}Learning rate                              & 2e-5  & 5e-4      & 2e-4   & 2e-4   & 5e-4                         & 1e-4     \\
\hspace{3mm}Epochs per update                          & 3     & 3         & 10     & 10     & 10                           & 30       \\
\hspace{3mm}Discriminator learning rate                & 1e-5  & 5e-5      & 5e-5   & 5e-4   & 5e-4                         & 5e-4     \\
\hspace{3mm}Discriminator batch size                   & 32    & 512       & 256    & 128    & 256                          & 256      \\
\hspace{3mm}Discriminator epochs per update            & 3     & 10        & 10     & 10     & 10                           & 10       \\
\hspace{3mm}Weighting coefficient $\beta$                                    & 1     & 1         & 0      & 1      & 0                            & 0        \\
\hspace{3mm}Mixing coefficient $\lambda_1$ (wPOfD)     & 0.9   & 0.01      & 0.9    & 0.01   & 0.5                          & 0.01     \\
\hspace{3mm}Annealing coef $\lambda_1^{sc}$ (wPOfD sc) & 1     & 5         & 2      & 0.5    & 0.5                          & 5        \\
\end{tabular}
\label{tab:hyperparameters}
\end{table*}

\begin{table*}[t]
\centering
\caption{Hyperparameters for all environments and algorithms (continued)}
\begin{tabular}{lcccccc}
\hline
Hyperparameter                                               & \multicolumn{1}{l}{Chain} & \multicolumn{1}{l}{Pommerman} & \multicolumn{1}{l}{Walker} & \multicolumn{1}{l}{Hopper} & \multicolumn{1}{l}{Cheetah} & \multicolumn{1}{l}{Cartpole}  \\ \hline
                                      &     & & & & \\
\hspace{3mm}\textbf{DQfD, DDPGfD}              &     & & & & \\
\hspace{3mm}Buffer size                                & 1e5   & 1e6       & 1e5    & 1e5    & 1e5                          & 1e5      \\
\hspace{3mm}Q-network / Actor learning rate            & 2e-4  & 1e-4      & 1e-4   & 2e-4   & 5e-4                         & 1e-3     \\
\hspace{3mm}Critic learning rate                       & -     & -         & 1e-3   & 1e-3   & 1e-3                         & 1e-3     \\
\hspace{3mm}Batch size                                 & 32    & 200       & 128    & 128    & 128                          & 128      \\
\hspace{3mm}Epochs per update                          & 1     & 1         & 1      & 1      & 1                            & 4        \\
\hspace{3mm}Demos priority bonus $\epsilon_{d}$        & 0.5   & 1         & 1      & 1      & 1                            & 0.2      \\
\hspace{3mm}Number of steps in n-step loss             & 10    & 10        & 20     & 10     & 5                            & 10       \\
\hspace{3mm}N-step loss weight $\lambda_{1}$           & 0.01  & 0.1       & 0.1    & 0.1    & 0.1                          & 0.1      \\
\hspace{3mm}Supervised loss weight $\lambda_{2}$       & 0.1   & 0.8       & -      & -      & -                            & -        \\
\hspace{3mm}l2 regularization weight                   & 1e-3  & 1e-4      & 1e-4   & 1e-4   & 1e-3                         & 1e-3     \\
                                      &     & & & & \\
\hspace{3mm}\textbf{TRPOfD}              &     & & & & \\
\hspace{3mm}Max KL $\delta$                            & 1e-4  & 2e-3      & 2e-3   & 2e-4   & 5e-3                         & 2e-4     \\
\hspace{3mm}Batch size                                 & 1000  & 8000      & 8000   & 8000   & 8000                         & 8000     \\
\hspace{3mm}Epochs per update                          & 1     & 1         & 1      & 1      & 1                            & 1        \\
\hspace{3mm}Max KL from expert $d_0$                   & 1e-3  & 1e-4      & 1e-4   & 1e-3   & 1e-4                         & 1e-3     \\
\hspace{3mm}Annealing coefficient $\epsilon$           & 5e-4  & 2e-3      & 5e-4   & 1e-3   & 2e-3                         & 2e-3     \\
\hspace{3mm}Recovery learning rate                     & 0.1   & 0.1       & 0.1    & 0.1    & 0.01                         & 0.01     \\
                                      &     & & & & \\
\hspace{3mm}\textbf{BC}              &     & & & & \\
\hspace{3mm}Learning rate                              & 1e-3  & 5e-4      & 2e-4   & 1e-4   & 5e-4                         & 5e-5     \\
\hspace{3mm}Batch size                                 & 32    & 256       & 512    & 256    & 256                          & 256      \\
\hspace{3mm}Number of epochs                           & 4096  & 1000      & 1000   & 1000   & 1000                         & 1000     \\
                                      &     & & & & \\
\hspace{3mm}\textbf{DT}              &     & & & & \\
\hspace{3mm}Learning rate                              & 1e-3  & 1e-4      & 1e-4   & 1e-4   & 1e-4                         & 1e-4     \\
\hspace{3mm}Batch size                                 & 32    & 64        & 64     & 64     & 64                           & 64       \\
\hspace{3mm}Number of batches                          & 20000 & 100000    & 100000 & 100000 & 100000                       & 100000   \\
\hspace{3mm}Number of attention layers                 & 1     & 3         & 3      & 3      & 3                            & 3        \\
\hspace{3mm}Dropout rate                               & 0     & 0.1       & 0.1    & 0.1    & 0.1                          & 0.1     
\end{tabular}
\label{tab:hyperparameters2}
\end{table*}


\end{appendix}

\end{document}